\newcommand{\numberQuestions}{246}
\newcommand{\numberEval}{5339}
\newcommand{\numberSlides}{21}
\newcommand{\numberStudents}{46}

\documentclass[runningheads]{llncs} % LLNCS macro package for Springer Computer Science proceedings

\usepackage[T1]{fontenc} % Other font encondings may result in incorrect characters
\usepackage{graphicx} % For displaying figures
\usepackage{color}
\usepackage{float}

\usepackage[margin=4.3cm]{geometry} % for ArXiv submission

\usepackage[protrusion]{microtype} % Nicer typography
\usepackage[breaklinks,colorlinks,linkcolor=blue,urlcolor=blue,citecolor=blue]{hyperref} % Links
\usepackage{tikz}
\newcommand\copyrighttext{%
  \scriptsize This preprint has not undergone peer review or any post-submission improvements or corrections. The Version of Record of this contribution is published in \textit{Technology Enhanced Learning for Inclusive and Equitable Quality Education, ECTEL 2024, Krems, Austria, September 16--20, 2024, Proceedings}, and is available online at \url{https://doi.org/10.1007/978-3-031-72312-4_18}.
  \smallskip
  
  \textcopyright\ 2024. Please cite this article as follows: I. Lodovico Molina, V. Švábenský, T. Minematsu, L.~Chen, F. Okubo, A. Shimada: \textit{Comparison of Large Language Models for Generating Contextually Relevant Questions}. In Proceedings of 19th European Conference on Technology Enhanced Learning (ECTEL), Springer, 2024. DOI: \href{https://doi.org/10.1007/978-3-031-72312-4_18}{10.1007/978-3-031-72312-4\_18}.}
\newcommand\copyrightnotice{%
\begin{tikzpicture}[remember picture,overlay]
\node[anchor=north,yshift=-24pt] at (current page.north) {\fbox{\parbox{\dimexpr\textwidth-\fboxsep-\fboxrule\relax}{\copyrighttext}}};
\end{tikzpicture}%
}

\begin{document}

\title{Comparison of Large Language Models for Generating Contextually Relevant Questions}

\titlerunning{Comparison of LLMs for Generating Contextually Relevant Questions}
% If the paper title is too long for the running head, set an abbreviated title

% \author{Anonymous Authors}
\author{
Ivo LODOVICO MOLINA\inst{1} \and
Valdemar ŠVÁBENSKÝ\inst{1}\orcidID{0000-0001-8546-280X}\thanks{Corresponding author. \href{mailto:valdemar.research@gmail.com}{valdemar.research@gmail.com}} \and
Tsubasa MINEMATSU\inst{2} \and
Li CHEN\inst{1} \and
Fumiya OKUBO\inst{1} \and
Atsushi SHIMADA\inst{1}
}

\authorrunning{I. Lodovico Molina et al.}
% First names are abbreviated in the running head.
% If there are more than two authors, 'et al.' is used.

% \institute{Anonymous Institutions}
\institute{Faculty of Information Science and Electrical Engineering, Kyushu University, Japan \and
Data-Driven Innovation Initiative, Kyushu University, Japan}

\maketitle

\begin{abstract}
This study explores the effectiveness of Large Language Models (LLMs) for Automatic Question Generation in educational settings. Three LLMs are compared in their ability to create questions from university slide text without fine-tuning. Questions were obtained in a two-step pipeline: first, answer phrases were extracted from slides using Llama 2-Chat 13B; then, the three models generated questions for each answer. To analyze whether the questions would be suitable in educational applications for students, a survey was conducted with $\numberStudents$ students who evaluated a total of $\numberQuestions$ questions across five metrics: clarity, relevance, difficulty, slide relation, and question-answer alignment. Results indicate that GPT-3.5 and Llama 2-Chat 13B outperform Flan T5 XXL by a small margin, particularly in terms of clarity and question-answer alignment. GPT-3.5 especially excels at tailoring questions to match the input answers. The contribution of this research is the analysis of the capacity of LLMs for Automatic Question Generation in education.

\keywords{Generative AI \and Question Generation \and AI in Education.}

\copyrightnotice
\end{abstract}

\section{Introduction}
\label{Introduction}

The rapid growth of Large Language Models (LLMs) has revolutionized numerous sectors, including education \cite{olga2023generative}. LLMs can focus on and transform relevant parts of a given text, making them suitable also for Automatic Question Generation (QG). Generating educational questions is beneficial for teachers and students, as it supports comprehension, skill development, and assessment \cite{barak1996review}, \cite{kurdi2019systematic}. Moreover, automating QG provides scalability and personalizes the learning path.

Providing an appropriate context for QG is crucial in order to produce questions that are relevant to the educational material. Before LLMs appeared, related work~\cite{xia2023improving}, \cite{liu2023qgae}, \cite{Wang_Wang_Tao_Zhang_Xu_2020} implemented context-specific QG models using well-known datasets. Other work focused on educational environments, such as in \cite{kurdi2019systematic}, where resources like school repositories, Wikipedia, or other websites provided context. However, the limitation of the past work is that the context was difficult to provide and remained static once set. With the recent advancements, LLMs can facilitate more flexible QG that is \textit{contextually relevant}.

In university education, slide-based teaching materials are prevalent, and slide text contains valuable content usable for QG. Slide-based QG can be beneficial in numerous ways, for example, if a student has only the slide material available and needs to challenge their knowledge and understanding using an automated tool. In another case, an instructor can automatically generate stimulating questions based on their slides without any intermediate steps, emphasizing what they consider important for the subject. Depending on the purpose of the generated questions, a lot of previous work has been reviewed in~\cite{kurdi2019systematic}. However, no works were found using LLMs with slide text as context.

This paper evaluates and compares the effectiveness of different LLMs in generating contextually relevant educational questions, using only slide text as context. We investigate the advantages and weaknesses of deploying three advanced LLMs: GPT-3.5 Turbo \cite{brown2020language} by OpenAI, Flan T5 XXL \cite{chung2022scaling}, and Llama 2-Chat 13B~\cite{touvron2023llama}. For this purpose, we create the slide-based automatic question generator and subsequently analyze the quality of the generated questions.

The challenge in utilizing slide text arises from the complexity of extracting meaningful answer phrases. The grammar, structure, and the amount of information in the slides can vary from slide to slide.

This paper focuses on the student perspective, because in our use case, students are the intended users of the generated questions. For example, students may use the questions for their knowledge reinforcement or diagnostic quizzes. Therefore, we pose the following research question: \textit{How do the three LLMs differ in terms of student evaluations of the generated questions?}

\section{Research Methods}
\label{proposed_method}

\subsection{Answer Extraction/Generation (Step 1)}
\label{subsec:answer-extraction}

A crucial step in our method is to obtain relevant concepts (called ``answers'' in this work) that are used to guide the QG later. From the three LLMs, only Llama 2-Chat 13B was always used for the answer generation. Its performance was very good: better than Flan T5 and about the same as GPT 3.5, which is, however, not free, so we decided to use the open model. The input for the LLM in this step is the corresponding \textit{prompt} for this task and the \textit{context}, which is the slide text from which the answer was extracted. It is important to separate the text used to obtain the answers, and the context that only gives more information to the model. The final output is a list of LLM-generated answers.

\subsection{Question Generation by the LLMs (Step 2)}

All three compared models (GPT-3.5 Turbo, Flan T5 XXL, and Llama 2-Chat 13B) were used for this QG step. The input is the corresponding \textit{prompt} for this task, the \textit{answer} generated in the previous step (only one from the list) that contains the meaningful information, and the \textit{context} (which is the same as in step 1). The final output is a single question generated by the LLM.

The three LLMs were chosen based on their general performance in various Natural Language Processing (NLP) tasks and availability. GPT-3.5 Turbo is a closed model by OpenAI, which can be accessed using APIs. Since it is widely used as a reference, we included it for that purpose as well. The other two models are open and have similar amount of parameters, but differ in architectures and pre-training, which is why it can be interesting to compare them. 

\subsection{Question Evaluation}
\label{eval}

For this study, we generated a total of $\numberQuestions$ questions from $\numberSlides$ different slides from the course Pattern Recognition taught at Kyushu University. These slides were selected from 7 different units of the course because they had a sufficient amount of meaningful content, especially including the crucial concepts in the subject. The slides explain topics related to machine learning (in English language).

We sought to understand different aspects of the generated questions, the strengths and weaknesses of the LLMs, and the models' behavioral scenarios that commonly manifested during the QG. To achieve this, we examined five features of a question, which were based on the metrics defined in \cite{elkins2023useful}.

\begin{itemize}
    \item \textit{Clarity}: evaluates how precise and easily understandable a question is. It should be clear and unambiguous to prevent student confusion.
    \item \textit{Relevance}: assesses how closely a question-answer pair is tied to the subject unit's core topic. It measures how well a question-answer pair contributes to understanding the subject matter.
    \item \textit{Difficulty}: evaluates how challenging it is to answer a question and how explicit the answer is in the context.
    \item \textit{Slide relation}: evaluates how well a question-answer pair aligns with the content presented in the one specifically provided slide.
    \item \textit{Question-answer (QA) alignment}: evaluates how accurate the provided answer is in relation to the question.
\end{itemize}

To analyze the generated questions, we conducted a survey with $\numberStudents$ undergraduate and graduate students of computer science degree programs at Kyushu University. The study participants were self-selected; they responded to a call for taking part in the research. There was reward for participation (gift cards).

To conduct the survey, we developed a custom web application in which the students can see the slides from the dataset in order to better understand the subject. The $\numberSlides$ slides were randomly distributed to the students to maintain a similar amount of evaluations for each slide. Some students evaluated more than one slide because not all the slides have the same amount of text, and the amount of extracted answers may differ as well. 

During the evaluation, the generated question and answer are presented, along with the five features and a scale from 1 (lowest score) to 5 (highest score) to rate them. In addition, the students marked their level of confidence (also from 1 to 5) to indicate if they are not confident or very confident respectively, in their evaluation of each question.

\section{Results and Discussion}
\label{results}

A total of $\numberQuestions$ different questions were rated by students, yielding $\numberEval$ evaluations. Between 18 and 24 students evaluated each question. The confidence level used for filtering the evaluations considerably modifies each distribution. When only the evaluations with higher confidence are analyzed, the distributions peak near the low or high ends of the scale. This indicates that the students felt more confident about the evaluations when the question features were more evident to them. Moreover, with increasing confidence thresholds, the distributions across different metrics become very similar to each other. 

Whether the distributions differ was statistically analyzed using Kruskal-Wallis H-tests, with a p-value cut-off of 0.05. Tables \ref{tab:kruskal_text_p_values}, \ref{tab:kruskal_text_p_values2}, and \ref{tab:kruskal_text_p_values3} show the test results separated by students' confidence. Significant values are highlighted.

\begin{table}[!ht]
\setlength{\tabcolsep}{3pt}
\renewcommand{\arraystretch}{1.1}
\centering

\caption{The p-values obtained from Kruskal-Wallis H-test between each pair of model's distributions for evaluations with confidence $\geq$ 3. Sample size: Llama2: 1517 evaluations; GPT3.5: 1533 evaluations; FlanT5: 1536 evaluations.}
\label{tab:kruskal_text_p_values}
%\resizebox{1.0\textwidth}{!}{
\begin{tabular}{l|ccccc}
\textbf{Model pair} & \textbf{Clarity} & \textbf{Relevance} & \textbf{Difficulty} & \textbf{Slide relation} & \textbf{QA align.} \\ \hline
Llama2 : GPT3.5     &          0.2194  & \textbf{0.0023}    & \textbf{0.0005}     & \textbf{0.0030}        & \textbf{0} \\
GPT3.5 : FlanT5     & \textbf{0.0002}  & \textbf{0}         & \textbf{0}          & \textbf{0}             & \textbf{0} \\
FlanT5 : Llama2     & \textbf{0.0096}  & 0.2482             & \textbf{0.0003}     & \textbf{0.0141}        & \textbf{0} \\
\end{tabular}
\\[3mm]

\caption{The p-values obtained from Kruskal-Wallis H-test between each pair of model's distributions for evaluations with confidence $\geq$ 4. Sample size: Llama2: 1128 evaluations; GPT3.5: 1172 evaluations; FlanT5: 1126 evaluations.}
\label{tab:kruskal_text_p_values2}
\begin{tabular}{l|ccccc}
\textbf{Model pair} & \textbf{Clarity} & \textbf{Relevance} & \textbf{Difficulty} & \textbf{Slide relation} & \textbf{QA align.} \\ \hline
Llama2 : GPT3.5     &           0.6912 & \textbf{0.0098}    & \textbf{0.0117}     &        \textbf{0.0124} &            \textbf{0} \\
GPT3.5 : FlanT5     &           0.0662 & \textbf{0.0006}    & \textbf{0}          &        \textbf{0}      &            \textbf{0} \\
FlanT5 : Llama2     &           0.1548 &          0.3614    & \textbf{0.0060}     &        \textbf{0.0334} &            \textbf{0} \\
\end{tabular}
\\[3mm]

\caption{The p-values obtained from Kruskal-Wallis H-test between each pair of model's distributions for evaluations with confidence = 5. Sample size: Llama2: 436 evaluations; GPT3.5: 489 evaluations; FlanT5: 411 evaluations.}
\label{tab:kruskal_text_p_values3}
\begin{tabular}{l|ccccc}
\textbf{Model pair} & \textbf{Clarity} & \textbf{Relevance} & \textbf{Difficulty} & \textbf{Slide relation} & \textbf{QA align.} \\ \hline
Llama2 : GPT3.5     &           0.6148 &             0.8879 &              0.2942 &                 0.0690 &       \textbf{0.0018} \\
GPT3.5 : FlanT5     &           0.2817 &             0.3366 &              0.0904 &        \textbf{0.0120} &       \textbf{     0} \\
FlanT5 : Llama2     &           0.5675 &             0.4223 &              0.5548 &                 0.4626 &       \textbf{0.0004} \\
\end{tabular}
\end{table}

\vspace{-1mm}

By keeping only the most confident evaluations (confidence = 5), more than 70\% of other evaluations are removed, which biases the results. Thus, for further analysis, the selected level of confidence for the evaluations is $\geq$ 3. This value has a good trade-off between the confidence and the bias that might manifest.

Due to the page limit restrictions, detailed results and supplementary materials are available at \url{https://github.com/limu-research/2024-ectel-qg}.

\subsection{Analysis of the Five Individual Question Features}

Overall, the student ratings for the five evaluated metrics (clarity, relevance, difficulty, slide relation, and QA alignment) were very good.
\begin{enumerate}
    \item Concerning the generation of clear questions, both Llama 2-Chat 13B and GPT 3.5 Turbo performed well, with 80\% of questions rated 4 or above. The Flan T5 XXL model encountered slightly more challenges, as some of its questions were rated lower for clarity by the students.
    \item In terms of relevance, all models had similar results. However, GPT 3.5 Turbo was slightly better.
    \item No major differences between the models were observed for question difficulty. GPT 3.5 Turbo tends to generate slightly easier questions, while Flan T5 XXL leans towards more difficult ones, making the overall range quite extensive.
    \item All LLMs scored similarly on slide relation with analogous results to the first two features; most output questions appropriately related to the slide content. However, GPT 3.5 Turbo again showed slightly better results.
    \item Interestingly, when it comes to the consonance between the generated questions and the provided answers, differences were more evident. Questions by GPT 3.5 Turbo aligned better with the answers compared to Llama 2-Chat 13B, and Flan T5 XXL presented the most issues in this aspect.
\end{enumerate}

\subsection{Benefits and Limitations for Educational Applications}

It is remarkable that even without fine-tuning, the models mostly scored high in the evaluations of the analyzed metrics. This indicates that the LLMs can be used for a satisfactory QG quickly and without specific knowledge about technical aspects of the models or specific training.

However, the LLMs still have a few weaknesses, mainly in ensuring the alignment between the output questions and the provided answers, and also in limiting bias towards certain types of questions (e.g., avoiding generic ``what is'' questions). Another constrain is that if the LLM output was to be integrated to educational software for practical deployment, so that the instructors would not have to interact with the LLM directly, the best-performing LLM is not freely available. Thus, the associated costs might limit the scalability of the solution.

Despite these limitations, the evaluated LLMs are advisable for educational applications, especially the support of students' personalized learning. This includes knowledge reinforcement or quick diagnostic quizzes, where occasional problems related to clarity or alignment can be tolerated, and the variance in difficulty is beneficial. In other tasks, such as automatic generation of formal tests and exams, the alignment and clarity of the question have to be precise, so the evaluated LLMs might not be suitable. 

Future work may optimize models and methodologies to further enhance the quality of AI-generated questions. For example, even though the LLMs performed well without fine-tuning, it would be interesting to explore whether fine-tuning would significantly improve the results. Related to this, future work can compare different prompts and describe which of them lead to better or worse questions. 

\section{Conclusion}
\label{Conclusion}

This research evaluated three LLMs in facilitating QG for educational applications. We aimed to gain deeper understanding of how Generative AI can be used to produce contextually relevant questions using slide text as context. The proposed pipeline to extract the concepts or ``answers'' resulted in a unique and effective method to process slides and obtain more granular information for the question generation task. This method can be applied for other text-based educational materials besides slides, such as textbooks and websites. The resulting questions may contribute to effective learning, inspire content creation, and allow for assessment and knowledge reinforcement.

All the evaluated LLMs demonstrated capability in QG, scoring high in clarity, relevance, and slide relation. The models performed well without fine-tuning, making them immediately applicable. Despite some limitations in answer alignment and occasional biases that complicate question interpretation, these models have potential for educational applications. Nevertheless, for tasks requiring high precision and QA alignment, fine-tuning and further improvements are necessary.

Materials are available at \url{https://github.com/limu-research/2024-ectel-qg}.

\begin{credits}
\subsubsection{\ackname}
This work was supported by JST CREST Grant Number JPMJCR22D1 and JSPS KAKENHI Grant Number JP22H00551, Japan.

\end{credits}

\bibliographystyle{splncs04}
\bibliography{references}

\end{document}